\pdfoutput=1

\documentclass[11pt]{article}

\usepackage{EMNLP2023}

\usepackage{times}
\usepackage{latexsym}
\usepackage{todonotes}
\usepackage{float}
\usepackage{graphicx}
\usepackage{caption}
\usepackage{subcaption}
\usepackage{tablefootnote}
\usepackage[normalem]{ulem}
\usepackage{amsmath}

\DeclareMathOperator*{\mean}{mean}
\DeclareMathOperator*{\var}{var}

\usepackage[T1]{fontenc}

\usepackage[utf8]{inputenc}

\usepackage{microtype}

\usepackage{inconsolata}

\title{DynaSemble: Dynamic Ensembling of Textual and Structure-Based Models for Knowledge Graph Completion}

\author{Ananjan Nandi \hskip 1em  
Navdeep Kaur  \hskip 1em
Parag Singla \hskip 1em Mausam \\
  Indian Institute of Technology, Delhi \\
  \texttt{\{tgk.ananjan, navdeepkjohal\}@gmail.com} \quad 
   \texttt{\{parags, mausam\}@cse.iitd.ac.in}  
}

\begin{document}
\maketitle
\begin{abstract}
We consider two popular approaches to Knowledge Graph Completion (KGC): textual models that rely on textual entity descriptions, and structure-based models that exploit the connectivity structure of the Knowledge Graph (KG). Preliminary experiments show that these approaches have complementary strengths: structure-based models perform exceptionally well when the gold answer is easily reachable from the query head in the KG, while textual models exploit descriptions to give good performance even when the gold answer is not easily reachable. In response, we propose \texttt{DynaSemble}, a novel method for learning query-dependent ensemble weights to combine these approaches by using the distributions of scores assigned by the models in the ensemble to all candidate entities. \texttt{DynaSemble} achieves state-of-the-art results on three standard KGC datasets, with up to 6.8 pt MRR and 8.3 pt Hits@1 gains over the best baseline model for the WN18RR dataset. 
\end{abstract}

\section{Introduction}

The task of Knowledge Graph Completion (KGC) can be described as inferring missing links in a Knowledge Graph (KG) based on given triples ($\fol{h}, \fol{r}, \fol{t}$), where $\fol{r}$ is a relation that exists between the head entity $\fol{h}$ and the tail entity $\fol{t}$. Several KGC approaches, such as NBFNet~\cite{NBFNet2021} and  RGHAT~\cite{RGHAT2020}, exploit the underlying graph structure, often using GNNs. 
On the other hand, textual models such as SimKGC~\cite{SimKGC2022} and HittER~\cite{chen-etal-2021-hitter} leverage pre-trained large language models (LLMs) such as BERT~\cite{BERT2019} to utilize textual descriptions of the KG entities and relations for KGC. 

Our preliminary experiments suggest that when the gold answer $\fol{t}$ for query ($\fol{h}, \fol{r}, ?$) is reachable from $\fol{h}$ via a path of reasonable length in the KG, structure-based models tend to outperform textual models. In contrast, textual models use textual descriptions to perform better than structure-based models when $\fol{t}$ is not easily reachable from $\fol{h}$. Motivated by our findings, we seek to explore how ensembling, an approach currently underrepresented in KGC literature (see \citet{mftfensemble} for an example), can effectively harness the complementary strengths of these models.

Consequently, we propose \texttt{DynaSemble}: a novel, simple, model-agnostic and lightweight method for learning ensemble weights such that the weights are (i) query-dependent and (ii) learned from statistical features obtained from the distribution of scores assigned by individual models to all candidate entities. This approach results in a new state-of-the-art baseline when applied on two strong KGC models: SimKGC and NBFNet, which are textual and structure-based in nature, respectively.

On three KGC datasets, we find that applying \texttt{DynaSemble} to SimKGC and NBFNet consistently improves KGC performance, outperforming best individual models by up to 6.8 pt MRR and 8.3 pt Hits@1 on the WN18RR dataset. To the best of our knowledge, our results are state of the art for all three datasets. Further experiments (including a fourth dataset to which NBFNet does not scale) show that \texttt{DynaSemble} generalises to ensembling with another KG embedding model, RotatE \cite{sun2018rotate}, with similar gains. We also demonstrate that \texttt{DynaSemble} outperforms conventional model-combination techniques such as static ensembling (where the ensemble weight is a tuned constant hyperparameter) and re-ranking. We release all code\footnote{\url{https://github.com/dair-iitd/KGC-Ensemble}} to guide future research.

\section{Background and Related Work}
\label{sec:background and related work}
\vspace{0.5ex}
 \noindent \textbf{Task:} We are given an incomplete KG $\mathcal{K} = (\mathcal{E}, \mathcal{R}, \mathcal{T})$ consisting of entities $\mathcal{E}$, relation set $\mathcal{R}$ and 
set of triples $\mathcal{T}=\{(\fol{h},\fol{r},\fol{t})\}$  (where $\fol{h}, \fol{t} \in \mathcal{E}$ and $\fol{r} \in \mathcal{R}$). The goal of KGC is to answer queries of the form  $(\fol{h},\fol{r},?)$ or $(?,\fol{r},\fol{t})$ to predict missing links, with corresponding answers $\fol{t}$ and $\fol{h}$. We model $(?,\fol{r},\fol{t})$ as $(\fol{t},\fol{r^{-1}},?)$ queries in this work.

\vspace{0.5ex}
\noindent \textbf{Overview of Related Work:} 
\label{sec:related work}
We focus on three types of KGC models. The first type consists of Graph Neural Network (GNN) based models such as NBFNet~\cite{NBFNet2021}, RGHAT~\cite{RGHAT2020} that leverage neighborhood information to train distinct GNN architectures. The second type contains textual models such as KG-BERT~\cite{KGBERT2020}, HittER~\cite{chen-etal-2021-hitter} and SimKGC~\cite{SimKGC2022} which fine-tune a pre-trained LLM on textual descriptions of entities and relations for KGC. The third type involves models such as RotatE~\cite{sun2018rotate} and ComplEx~\cite{complex2016, typecomplex} that learn low-dimensional embeddings for entities and relations and compose them by employing unique scoring functions. Unification of these approaches has not been extensively studied in KGC literature. VEM2L~\cite{vem2l} proposes a method to encourage multiple KGC models to learn from each other during training. KGT5~\cite{saxena2022sequence} finds that their textual model struggles when the query has a large number of correct answers in the training set and routes those queries to a structure-based model as a consequence, exhibiting some performance gains.
Since our main experiments are based on NBFNet, SimKGC, HittER and RotatE, we describe these next.


\vspace{0.5ex}
 \noindent \textbf{NBFNet:} Neural Bellman-Ford Network (NBFNet) is a path-based link prediction model that introduces neural functions into the Generalized Bellman-Ford (GBF) Algorithm~\cite{GBFAlgo2010}, which in turn models the path between two nodes in the KG through generalized sum and product operators. This formulates a novel GNN framework that learns entity representations for each candidate tail $\fol{t}$ conditioned on $\fol{h}$ and $\fol{r}$ for each query $(\fol{h},\fol{r},?)$. The score of any candidate $\fol{t}$ is then computed by applying an MLP to its embedding. 

\vspace{0.5ex}
 \noindent \textbf{SimKGC:} SimKGC is an LLM-based KGC model that employs a bi-encoder architecture to generate the score of a given triple ($\fol{h},\fol{r},\fol{t}$). The model considers two pre-trained BERT~\cite{BERT2019} models. The first model is finetuned on a concatenation of textual descriptions of $\fol{h}$ and $\fol{r}$ to generate their joint embedding $\textbf{e}_{hr}$ and the second model is finetuned on the textual description of $\fol{t}$ to generate the embedding $\textbf{e}_{t}$. The score for the triple is the cosine similarity between $\textbf{e}_{hr}$ and $\textbf{e}_{t}$. 

 \vspace{0.5ex}
 \noindent \textbf{HittER:} HittER proposes a hierarchical transformer-based approach for jointly learning entity and relation embeddings by aggregating information from the graph neighborhood. A transformer provides relation-dependent entity embeddings for the neighborhood of an entity, which are then aggregated by another transformer. These embeddings are trained using a joint masked entity prediction and link prediction task. 

\vspace{0.5ex}
 \noindent \textbf{RotatE:} 
RotatE is a KG Embedding model that maps entities and relations to a complex vector space and models each relation $\fol{r}$ as a complex rotation from the head $\fol{r}$ to the tail $\fol{t}$ for triple ($\fol{h},\fol{r},\fol{t}$). More specifically, the scoring function of RotatE is $ \left \Vert \textbf{h}\circ \textbf{r} -\textbf{t} \right\Vert$ where 
$\textbf{h}, \textbf{t} \in \mathbb{C}^{k}$ are the complex embedding of $\fol{h}$ and $\fol{t}$,
and $\circ$ is the Hadamard product. 
 
\section{DynaSemble}
\label{sec:main}
Our goal is to dynamically ensemble $k$ KGC models $\fol{M_i}$, which may be textual or structure-based, to maximize performance. Each model $\fol{M_i}$ assigns a score $\fol{M_i(h,r,t)}$ to all candidate tails $\fol{t} \in \mathcal{E}$ for query $\fol{q} = (\fol{h},\fol{r},?)$. These models are trained independently and their parameters are frozen before ensembling. We formulate the ensemble $\fol{E}$ as:
\vspace{-0.10in}
\begin{equation*}
\label{eq: Formulation}
\fol{E(h,r,t)} = \sum_{i=1}^{k}\fol{w_i(q)M_i(h,r,t)}
\end{equation*}

\noindent where $\fol{E(h,r,t)}$ is the ensemble score for $\fol{t}$ given query $\fol{q} = \fol{(h,r,?)}$. We first normalize these scores as described below.

\vspace{0.5ex}
\noindent \textbf{Normalization:} 
To bring the distribution of scores assigned by each model $\fol{M_i}$ over all $\fol{t} \in \mathcal{E}$ in the same range for each query, we max-min normalize the scores obtained from all models $\fol{M_i}$ separately: 
\begin{equation*}
\vspace{-0.05in}
\label{eq: Normalization1}
\fol{M_i(h,r,t)} \leftarrow \fol{M_i(h,r,t)} - \min_{\fol{t'} \in \, \mathcal{E}}{\fol{M_i(h,r,t')}}
\vspace{-0.02in}
\end{equation*}
\begin{equation*}
\vspace{-0.05in}
\label{eq: Normalization2}
\fol{M_i(h,r,t)} \leftarrow \frac{\fol{M_i(h,r,t)}}{\max_{t' \in \mathcal{E}}{\fol{M_i(h,r,t')}}}
\end{equation*}

The scores obtained after normalization lie in the range [0,1] for all models. We next describe the simple model used to learn the query-dependent ensemble weights $\fol{w_i}$. 

\begin{table*}[t]
 \caption{Results on four datasets for our baselines and approach. $\fol{[NBF]}$, $\fol{[Sim]}$, $\fol{[Hit]}$and $\fol{[RotE]}$ represent $\fol{NBFNet}$, $\fol{SimKGC}$, $\fol{HittER}$ and $\fol{RotatE}$ models. $\fol{[NBF]}$ does not scale up to YAGO3-10. We use model checkpoints published by the authors for $\fol{[Hit]}$ on the WN18RR and FB15k-237 datasets. Best individual model results are underlined. }
 \vspace{-0.07in}
 \label{tab:tablemainMRRAndHitsAt10}
 \centering
\small
\begin{center}
\begin{tabular}{|p{2.6cm}|p{0.6cm}p{0.5cm}p{0.75cm}|
p{0.6cm}p{0.5cm}p{0.75cm}|
p{0.6cm}p{0.5cm}p{0.75cm}|
p{0.6cm}p{0.5cm}p{0.75cm}|} 
\Xhline{3\arrayrulewidth} 
\multirow{2}{2em}{\textbf{Model}} & \multicolumn{3}{c|}{\textbf{WN18RR}} &
\multicolumn{3}{c|}{\textbf{FB15k-237}} &  \multicolumn{3}{c|}{\textbf{CoDex-M}} & \multicolumn{3}{c|}{\textbf{YAGO3-10}}\\
 & MRR & H@1 & H@10 & MRR & H@1 &  H@10 & MRR & H@1 &  H@10 & MRR & H@1 & H@10 \\
\Xhline{3\arrayrulewidth}
 $\fol{[Sim]}$ & \underline{66.4} & \underline{58.5} & \underline{80.3} & 32.1 &23.2 &50.5 & 29.1 &21.0 & 45.2 & 15.8 &10.0  & 27.3\\
  $\fol{[Hit]}$ & 50.3 & 46.3 & 58.5 & 37.2 &27.8 &55.8 & - & - & - & - &-  & -\\
$\fol{[NBF]}$ & 54.2 &48.6 & 65.7 & \underline{40.5} & \underline{31.0} & \underline{59.4} & \underline{35.3} & \underline{27.0}  & \underline{51.4} & - & - & -  \\
$\fol{[RotE]}$ &47.7 &43.9 &55.2 & 33.7 &24.0 & 53.2   &33.5  & 26.3  & 46.9 & \underline{49.3} & \underline{39.9} & \underline{67.1}\\
\Xhline{3\arrayrulewidth}
$\fol{[Sim]}$+$\fol{[NBF]}$ &\textbf{73.2}  &\textbf{66.9} &\textbf{85.7} & 42.7  & 33.2 & 61.5 & 38.9 & 30.5 & \textbf{54.8} & - & - & -\\
 $\fol{[Sim]}$+$\fol{[RotE]}$ &68.0 & 60.7& 80.7 & 36.6 & 26.9&56.3  &36.3  & 28.1 & 51.7 & \textbf{50.6}  & \textbf{41.3} & \textbf{67.9} \\
 $\fol{[Hit]}$+$\fol{[NBF]}$ & 56.8 & 51.7 & 67.1 & 42.1 &32.6 &60.8 & - & - & - & - &-  & -\\
 $\fol{[Hit]}$+$\fol{[RotE]}$ & 51.4 & 47.7 & 59.4 & 38.5 &29.0 &57.2 & - & - & - & - &-  & -\\
 \Xhline{3\arrayrulewidth}
$\fol{[Sim]}$+$\fol{[NBF]}$+$\fol{[RotE]}$ &\textbf{73.2}  &\textbf{66.9} &\textbf{85.7} & \textbf{43.0}  & \textbf{33.4} & \textbf{62.0} & \textbf{40.0} & \textbf{31.2} & \textbf{54.8} & - & - & -\\
$\fol{[Hit]}$+$\fol{[NBF]}$+$\fol{[RotE]}$ & 57.0 & 51.9 & 67.3 & 42.4 &32.8 &60.9 & - & - & - & - &-  & -\\
 \Xhline{3\arrayrulewidth}
\end{tabular}
\end{center}
\vspace{-1ex}
\end{table*}

\vspace{0.5ex}
\noindent \textbf{Model:}
We extract the following features from the score distribution of each model $\fol{M_i}$:
\begin{equation*}
\label{eq: Feat}
\fol{f(M_i, q)} = \fol{\mean_{t' \in \mathcal{E}}({M_i(h,r,t')})} || \fol{\var_{t' \in \mathcal{E}}({M_i(h,r,t')})}
\end{equation*}
In the above equations, $\mean()$ and $\var()$ are the standard mean and variance functions respectively, whose outputs are concatenated to obtain the feature. 
This choice is driven by the insight that the variance and mean of the distribution of scores computed by any model over $\mathcal{E}$ is correlated to the model confidence.
A more detailed discussion, along with an exploration of other possible feature sets can be found in Appendix \ref{appendix:featureselection}.

Next, we concatenate these features for all $\fol{M_i}$ to obtain a final feature vector that is passed to an independent 2-layer MLP ($\fol{MLP_i}$) for each model $\fol{M_i}$ to learn query-dependent $\fol{w_i}$:
\begin{equation*}
\label{eq: FeatConcat}
\fol{w_i(q)} = \fol{MLP_i}(\fol{f(M_1,q) || f(M_2,q) || ... || f(M_k,q)})
\end{equation*}

Intuitively, this concatenation informs each MLP about the relative confidence of all models regarding their predictions, enhancing the ensemble weight computation for corresponding models.
Note that our approach is agnostic to models $\fol{M_i}$. 


Our experiments in this paper mostly involve only one textual model. Therefore, we learn the ensemble weights for the other models with respect to this textual model, which is assigned a fixed weight of 1. This decreases the parameter count while still being as expressive as learning distinct ensemble weights for all models. The method for learning these other weights is unchanged.

\vspace{0.5ex}
\noindent \textbf{Loss Function:}
We train \texttt{DynaSemble} on the validation set (traditionally used to tune ensemble weights) using margin loss between the score of the gold entity and a set of negative samples. The train set is not used since all models are likely to give high-confidence predictions on its triples (Appendix \ref{appendix:choice}). If the gold entity is $\fol{t^{*}}$ and the set of negative samples is $\fol{N}$, the loss function $\mathcal{L}$ for query $\fol{q} = \fol{(h,r,?)}$ is:
\begin{equation*}
\label{eq: Loss}
\mathcal{L} = \sum_{\fol{t} \in \fol{N}}{\max{\fol{(E(h,r,t) - E(h,r,t^{*}) + m, 0)}}}
\end{equation*}

where $\fol{m}$ is the margin hyperparameter. This hyperparameter ensures that the generated ensemble weights stay numerically stable during training. In practice, we find that this loss function can be substituted for a cross-entropy loss as well.

\section{Experiments}
\label{sec:Experiments}
\noindent \textbf{Datasets:} We use four datasets for evaluation: WN18RR~\cite{WN18RRDataset2018}, FB15k-237~\cite{FB15K237dataset2015}, CoDex-M~\cite{CoDeX2020} and YAGO3-10~\cite{Yago3102015}.  For each triple in the test set, we answer queries $\fol{(h, r, ?)}$ and $\fol{(t, r^{-1}, ?)}$ with answers $\fol{t}$ and $\fol{h}$. We report the Mean Reciprocal Rank (MRR) and Hits@k (H@1, H@10) under the filtered measures~\cite{TransE2013}. Details and data statistics are in Appendix \ref{appendix: datastatistics}.

\vspace{0.5ex}
\noindent \textbf{Baselines:} We use $\fol{SimKGC}$ ($\fol{[Sim]}$ in tables) and $\fol{HittER}$ ($\fol{[Hit]}$ in tables) as strong textual model baselines. $\fol{NBFNet}$ ($\fol{[NBF]}$ in tables) serves as a strong structure-based model baseline. We also present results with $\fol{RotatE}$ ($\fol{[RotE]}$ in tables) to showcase the generalisation of our method to KG embedding models. We have reproduced the numbers published by the original authors for these baselines, and use model checkpoints published by the authors\footnote{\url{https://github.com/microsoft/HittER}} for $\fol{[Hit]}$ on the WN18RR and FB15k-237 datasets. Since $\fol{[NBF]}$ does not scale up to YAGO3-10 with reasonable hyperparameters on our hardware, we omit those results. We represent \texttt{DynaSemble} of models by $\fol{+}$ in tables. 

\vspace{0.5ex}
\noindent \textbf{Experimental Setup:} 
All baseline models are frozen after training using optimal configurations. Ensemble weights are trained on the validation split, using Adam as the optimizer with a learning rate of 5.0e-5. We use 10,000 negative samples per query. MLP hidden dimensions are set to 16 and 32 for ensemble of 2 and 3 models respectively. MLP weights are initialized uniformly in the range [0, 2]. \texttt{DynaSemble} training converges in a single epoch, making our method fast and efficient.    

\vspace{0.5ex}
\noindent \textbf{Results:} We report \texttt{DynaSemble} results  in Table \ref{tab:tablemainMRRAndHitsAt10} (more details in Appendix \ref{appendix:mainresultsindetail}). We observe a notable increase in performance after ensembling with $\fol{[Sim]}$ and $\fol{[Hit]}$ for both $\fol{[NBF]}$ and $\fol{[RotE]}$, which shows that our approach is performant for the ensembling of textual models with both structure-based and KG embedding models. In particular, we obtain 6.8 pt MRR and 8.4 pt Hits@1 improvement with $\fol{[Sim]} + \fol{[NBF]}$ over $\fol{[Sim]}$ on WN18RR. Ensembling with $\fol{[Sim]}$ results in substantial performance gains even when it is outperformed by structure-based models (on FB15k-237, CoDex-M and YAGO3-10 datasets). 

We find that ensembling of $\fol{[NBF]}$ and $\fol{[RotE]}$ with $\fol{[Sim]}$ results in larger improvements than with $\fol{[Hit]}$ (notably with a 16.4 pt MRR and 15.2 pt Hits@1 gap between $\fol{[Sim]} + \fol{[NBF]}$ and $\fol{[Hit]} + \fol{[NBF]}$). Even on the FB15k-237 dataset, where $\fol{[Hit]}$ outperforms $\fol{[Sim]}$ by 5.1 pt MRR and 4.6 pt Hits@1, $\fol{[Sim]} + \fol{[NBF]}$ narrowly outperforms $\fol{[Hit]} + \fol{[NBF]}$ by 0.6 pt MRR and 0.6 pt Hits@1. These observations suggest that $\fol{[Sim]}$ leverages the textual information in the knowledge graph more effectively than $\fol{[Hit]}$, thus acting as a better complement to the structure-based models.

On YAGO3-10, where $\fol{[RotE]}$ outperforms $\fol{[Sim]}$ by 33.5 pt MRR and 29.9 pt Hits@1, we still obtain 1.3 pt MRR and 1.4 pt Hits@1 gain with $\fol{[Sim]}$ + $\fol{[RotE]}$ over $\fol{[RotE]}$. Results for $\fol{[Sim]} + \fol{[NBF]}$ + $\fol{[RotE]}$ show that ensembling with $\fol{[RotE]}$ results in marginal gains over $\fol{[Sim]} + \fol{[NBF]}$, obtaining up to 1.1 pt MRR and 0.7 pt Hits@1 gain on CoDex-M. We hypothesize that the gains are marginal due to $\fol{[RotE]}$'s ability to implicitly capture and exploit structural information (explored in more detail in Appendix \ref{appendix:detailedablation} and \ref{appendix:rotatestructure}), making it somewhat redundant in the presence of $\fol{[NBF]}$. To the best of our knowledge, our best results on the WN18RR, FB15k-237 and CoDex-M datasets are state-of-the-art.

\section{Analysis}
\label{sec:analysis}
We perform four further analyses to answer the following questions: \textbf{Q1}. How does the behavior of textual and structure-based models vary with reachability? \textbf{Q2.} Do the weights learned by \texttt{DynaSemble} follow expected trends with reachability? \textbf{Q3.} Does \texttt{DynaSemble} improve performance over conventional model-combination techniques? \textbf{Q4.} How does \texttt{DynaSemble} of a textual and structure-based model compare to \texttt{DynaSemble} of two textual or structure-based models?

\vspace{0.5ex}
\noindent \textbf{Reachability Ablation:} To answer \textbf{Q1}, we divide the test set for each dataset into `reachable' and `unreachable' splits. A triple $\fol{(h,r,t)}$ is part of the reachable split if $\fol{t}$ can be reached from $\fol{h}$ with a path of length at most $\fol{l}$ (= 2) in the KG. If not, it is put in the unreachable split. We present split-wise results for $\fol{[NBF]}$, $\fol{[Sim]}$ and $\fol{[Sim]}$+$\fol{[NBF]}$ on the WN18RR and FB15k-237 datasets in Table \ref{tab:ReachabilityAblation}.

\begin{table}[th]
 \caption{Results on Reachable and Unreachable Split of $\fol{[NBF]}$, $\fol{[Sim]}$ and $\fol{[Sim]}$ + $\fol{[NBF]}$ on WN18RR and FB15k-237. Best individual model results are underlined.}
  \vspace{-0.07in}
 \label{tab:ReachabilityAblation}
 \centering
\small
\begin{center} 
\resizebox{\columnwidth}{!}{
\begin{tabular}{|p{1.5cm}|c|ccc|ccc|} 
\Xhline{3\arrayrulewidth}
\multirow{2}{4em}{\textbf{Dataset}} & \multirow{2}{4em}{\textbf{Model}} & 
\multicolumn{3}{c|}{\textbf{Reachable Split}} &
\multicolumn{3}{c|}{\textbf{Unreachable Split}} \\
& & MRR & H@1 & H@10 & MRR & H@1 & H@10 \\
\Xhline{3\arrayrulewidth}
 \multirow{3}{4em}{\textbf{WN18RR}}& $\fol{[NBF]}$  & \underline{89.7} & \underline{86.8} & \underline{95.7} & 26.0 & 18.3 & 41.8\\
 & $\fol{[Sim]}$ & 85.3 & 79.4 & 94.5 & \underline{51.8} & \underline{42.3} & \underline{69.0}\\
 & $\fol{[Sim]}$+$\fol{[NBF]}$ & \textbf{93.9} & \textbf{91.7} & \textbf{97.4} & \textbf{56.8} & \textbf{47.0} &  \textbf{76.4} \\
\cline{1-8}
 \multirow{3}{4em}{\textbf{FB15k}$-$\textbf{237}}& $\fol{[NBF]}$ & \underline{44.8} & \underline{35.2} & \underline{64.0} & 28.2 & 19.3 & 46.2\\
 & $\fol{[Sim]}$  & 31.5 & 22.6 & 49.6 & \underline{30.0} & \underline{21.2} & \underline{48.2}\\
 & $\fol{[Sim]}$+$\fol{[NBF]}$ & \textbf{46.5} & \textbf{36.8} & \textbf{65.3} & \textbf{32.3} & \textbf{23.1} & \textbf{50.6}\\
\Xhline{3\arrayrulewidth}
\end{tabular}
}
\end{center}
\vspace{-0.15in}
\end{table}

We observe that $\fol{[Sim]}$ outperforms $\fol{[NBF]}$ on the unreachable split (by up to 25.8 pt MRR and 24.0 pt Hits@1 for WN18RR), while $\fol{[NBF]}$ outperforms $\fol{[Sim]}$ on the reachable split (by up to 13.3 pt MRR and 12.6 pt Hits@1 for FB15k-237). This is because $\fol{[NBF]}$ can easily exploit knowledge of the KG structure to perform well on the reachable split, while $\fol{[Sim]}$ can instead use BERT to leverage textual descriptions to perform better on the unreachable split. The performance gap between $\fol{[Sim]}$ and $\fol{[NBF]}$ on the unreachable split is notably larger for WN18RR than for FB15k-237, which can be attributed to the sparsity of the WN18RR dataset, the unreachable split for which also has several entities unseen in the training data. In such cases, $\fol{[Sim]}$ achieves reasonable performance, whereas $\fol{[NBF]}$ lacks any paths for reasoning. Our ensemble obtains substantial gains over best individual models on both splits, with 4.2 pt MRR and 4.9 pt Hits@1 gain on the reachable split and 5.0 pt MRR and 4.7 pt Hits@1 gain on the unreachable split for WN18RR. More details in Appendix \ref{appendix:detailedablation}. 

\vspace{0.5ex}
\noindent \textbf{Analysis of Ensemble Weights:}
To answer \textbf{Q2}, we study the mean of the ensemble weight $\fol{w_2}$ for $\fol{[Sim]}$ + $\fol{[NBF]}$ over the queries in the reachable and unreachable splits of the datasets we use.  
We observe that this mean is consistently larger (by a margin of up to 17\% for WN18RR) on the reachable split than the unreachable split. This is because $\fol{[NBF]}$ tends to give better performance on the reachable split, and a larger $\fol{w_2}$ gives it more importance in the ensemble. More details and numbers are in Appendix \ref{appendix:detailedweights}, including results analyzing the non-trivial standard deviation of $\fol{w_2}$.
 
\vspace{0.5ex}
\noindent \textbf{Comparison with Conventional Techniques:}
To answer \textbf{Q3}, we present results for static ensembling and re-ranking using $\fol{[Sim]}$ and $\fol{[NBF]}$ for WN18RR and FB15k-237 datasets in Table \ref{tab:LearnedvsTunedWeights}. `Static ensembling' involves manually tuning the ensemble weight as a constant on the validation set. For $\fol{[NBF]}$-$\fol{[Sim]}$ re-ranking~\cite{TFRANKING2020LEARNINGTORANKWB}, we consider the top 100 entities by score from $\fol{[NBF]}$ for each query and re-rank them according to their $\fol{[Sim]}$ score. The rest of the entities are ranked according to $\fol{[NBF]}$. We present results for $\fol{[Sim]}$-$\fol{[NBF]}$ re-ranking as well for comparison. We also include results for the ensembling heuristic used in KGT5~\cite{saxena2022sequence} ($\fol{KGT5}$ Ensemble), which uses the textual model to answer queries that have no answers in the training set and the structure-based model to answer all other queries. 

\begin{table}[H]
 \caption{Comparison of Static, $\fol{KGT5}$ and Dynamic Ensembling and Re-ranking. $\fol{[X]}$-$\fol{[Y]}$ re-ranking indicates re-ranking of top 100 predictions from $\fol{[X]}$ using $\fol{[Y]}$.}
  \vspace{-0.07in}
 \label{tab:LearnedvsTunedWeights}
 \centering
\small
\begin{center} 
\resizebox{\columnwidth}{!}{
\begin{tabular}{|p{1.5cm}|c|ccc|} 
\Xhline{3\arrayrulewidth}
\textbf{Dataset} & \textbf{Approach} & MRR & H@1 & H@10 \\
\cline{1-5}
 \multirow{4}{4em}{\textbf{WN18RR}}& $\fol{[NBF]}$-$\fol{[Sim]}$ Re-rank & 63.5 & 57.1 & 74.9 \\
 & $\fol{[Sim]}$-$\fol{[NBF]}$ Re-rank& 60.7 & 53.3 & 76.0 \\
 & Static Ensemble & 72.2 & 65.5 & 85.4  \\
 & $\fol{KGT5}$ Ensemble & 66.6 & 58.7 & 80.3  \\
 & Dynamic Ensemble & \textbf{73.2} & \textbf{66.9} & \textbf{85.7}  \\
\cline{1-5}
 \multirow{4}{4em}{\textbf{FB15k}$-$\textbf{237}}& $\fol{[NBF]}$-$\fol{[Sim]}$ Re-rank & 32.7 & 23.3 & 52.5 \\
 & $\fol{[Sim]}$-$\fol{[NBF]}$ Re-rank& 38.9 & 30.0 & 56.5 \\
 & Static Ensemble & 41.9 & 32.7 & 60.1  \\
 & $\fol{KGT5}$ Ensemble & 31.1 & 22.3 & 49.3  \\
 & Dynamic Ensemble & \textbf{42.7} & \textbf{33.2} & \textbf{61.5}  \\
\cline{1-5}
\end{tabular}
}
\end{center}
\vspace{-0.15in}
\end{table}

We find that \texttt{DynaSemble} outperforms re-ranking, `$\fol{KGT5}$ ensembling' and `static ensembling' across datasets. Notably, \texttt{DynaSemble} beats re-ranking by 9.7 pt MRR and 9.8 pt Hits@1, $\fol{KGT5}$ ensembling by 5.6 pt MRR and 8.2 pt Hits@1, and static ensembling by 1.0 pt MRR and 1.4 pt Hits@1 on the WN18RR dataset. This highlights the utility of \texttt{DynaSemble} in comparison to existing model combination heuristics. We also perform a paired student's t-test to validate the statistical significance of the gains obtained from \texttt{DynaSemble} over "static ensembling", resulting in a t-value of 8.9 ($p < 0.001$) for the WN18RR dataset and 6.7 ($p < 0.01$) for the CoDex-M dataset. Further details can be found in Appendix \ref{appendix:sig}.

\vspace{0.5ex}
\noindent \textbf{Impact of Types of Ensembled Models:} To answer \textbf{Q4}, we contrast results for $\fol{[Sim]} + \fol{[NBF]}$ (\texttt{DynaSemble} of a textual and structure-based model) against $\fol{[Sim]} + \fol{[Hit]}$ (\texttt{DynaSemble} of two textual models) and $\fol{[NBF]} + \fol{[RotE]}$ (\texttt{DynaSemble} of two structure-based models) for WN18RR and FB15k-237 datasets in Table \ref{tab:ModelComparison}. 

\begin{table}[th]
 \caption{Results for $\fol{[Sim]}$ + $\fol{[NBF]}$, $\fol{[Sim]}$ + $\fol{[Hit]}$ and $\fol{[NBF]}$ + $\fol{[RotE]}$ on WN18RR and FB15k-237. Best individual model results are underlined.}
  \vspace{-0.07in}
 \label{tab:ModelComparison}
 \centering
\small
\begin{center} 
\resizebox{\columnwidth}{!}{
\begin{tabular}{|c|ccc|ccc|} 
\Xhline{3\arrayrulewidth}
\multirow{2}{4em}{\textbf{Model}} & 
\multicolumn{3}{c|}{\textbf{WN18RR}} &
\multicolumn{3}{c|}{\textbf{FB15k-237}} \\
 & MRR & H@1 & H@10 & MRR & H@1 & H@10 \\
\Xhline{3\arrayrulewidth}
  $\fol{[Sim]}$  & \underline{66.4} & \underline{58.5} & \underline{80.3} & 32.1 & 23.2 & 50.5\\
  $\fol{[Hit]}$  & 50.3 & 46.3 & 58.5 & 37.2 & 27.8 & 55.8\\
  $\fol{[NBF]}$  & 54.2 & 48.6 & 65.7 & \underline{40.5} & \underline{31.0} & \underline{59.4}\\
  $\fol{[RotE]}$  & 47.7 & 43.9 & 55.2 & 33.7 & 24.0 & 53.2\\
  \Xhline{3\arrayrulewidth}
  $\fol{[Sim]} + \fol{[NBF]}$ & \textbf{73.2} & \textbf{66.9} & \textbf{85.7} & \textbf{42.7} & \textbf{33.2} & \textbf{61.5} \\
  $\fol{[Sim]} + \fol{[Hit]}$ & 68.1 & 61.2 & 80.9 & 37.8 & 28.2 & 56.8 \\
  $\fol{[NBF]} + \fol{[RotE]}$ & 55.4 & 50.0 & 66.3 & 42.3 & 32.8 & 61.3 \\
\Xhline{3\arrayrulewidth}
\end{tabular}
}
\end{center}
\vspace{-0.15in}
\end{table}

\texttt{DynaSemble} achieves 1.7 pt MRR and 1.7 pt Hits@1 improvements over best individual models ($\fol{[Sim]}$) for $\fol{[Sim]} + \fol{[Hit]}$ on the WN18RR dataset and 1.8 pt MRR and 1.8 pt Hits@1 improvements over best individual models ($\fol{[NBF]}$) for $\fol{[NBF]} + \fol{[RotE]}$ on the FB15k-237 dataset, showing that \texttt{DynaSemble} generalizes to these settings. We further note that $\fol{[Sim]} + \fol{[NBF]}$ outperforms $\fol{[Sim]} + \fol{[Hit]}$ by 5.1 pt MRR and 5.7 pt Hits@1 and $\fol{[NBF]} + \fol{[RotE]}$ by 17.8 pt MRR and 16.9 pt Hits@1 on the WN18RR dataset. This trend persists for the FB15k-237 dataset, where $\fol{[Sim]} + \fol{[NBF]}$ marginally outperforms $\fol{[NBF]} + \fol{[RotE]}$ despite $\fol{[RotE]}$ outperforming $\fol{[Sim]}$ by 1.6 pt MRR and 0.8 pt Hits@1 individually. These observations are in line with our insights regarding the complementary strengths of textual and structure-based KGC approaches, which results in larger gains when models corresponding to different approaches are ensembled. 

\section{Conclusion and Future Work}
\label{sec:conclusion}
We present \texttt{DynaSemble}: a simple, novel, model-agnostic and lightweight dynamic ensembling approach for KGC, while also highlighting the complementary strengths of textual and structure-based KGC models. Our state-of-the-art results for a \texttt{DynaSemble} of SimKGC and NBFNet over three standard KGC datasets (WN18RR, FB15k-237 and CoDex-M) creates a new competitive ensemble baseline for the task. 
We release all code for future research. Future work includes tighter training-time unification methods, and
extensions to temporal \cite{timeplex, neustip} and multilingual KGC models \cite{multilingualkgc}.

\section*{Limitations}
We do not consider Neuro-Symbolic KGC approaches in this work, which have also recently started to give competitive results with other KGC approaches, through models such as RNNLogic \cite{RNNLogic2021} and extensions \cite{nandi23}. Our experiments consider ensembling of one textual model with multiple structural models. This is because most textual models in recent KGC literature are not competitive with SimKGC \cite{SimKGC2022}, therefore we do not expect large gains by including them along with SimKGC in an ensemble. The ensembling of multiple textual models with multiple structure-based models would be a possible future work. In models with substantial validation splits, learning query embeddings to augment the features we use to compute ensemble weights is also a possibility. 

\section*{Ethics Statement}
 We anticipate no substantial ethical issues arising due to our work on ensembling textual and structure-based models for KGC. Our work relies on other baseline models for ensembling. This may propagate any bias present in these baseline models, however ensembling may also reduce these biases. 

\section*{Acknowledgements}
This work is supported by IBM AI Horizons Network, grants from Google, Verisk, and Huawei, and the Jai Gupta chair fellowship by IIT Delhi. We thank the IIT-D HPC facility for its computational resources.

\bibliography{anthology,custom}
\bibliographystyle{acl_natbib}

\appendix

\begin{table*} 
\centering
\small
\caption{Statistics of Knowledge Graph datasets}
\vspace{-0.05in}
\begin{tabular}{|cccccc|}
\Xhline{3\arrayrulewidth}
Datasets & \#Entities & \#Relations & \#Training & \#Validation & \#Test \\
\hline
FB15k-237 & 14541 & 237 & 272,115 & 17,535 & 20,446 \\
WN18RR & 40,943 & 11 & 86,835 & 3,034 & 3,134 \\
Yago3-10 & 123182 & 36 & 1,079,040 & 5000 & 5000 \\
CoDex-M & 17050 & 71 & 185584 & 10310 & 10311  \\
\Xhline{3\arrayrulewidth}
\end{tabular}
\label{tab:KG statistics}
\vspace{0.12in}
\centering
\small
\caption{Results of on four datasets: WN18RR, FB15k-237, Yago3-10 and CoDex-M with ensemble of textual and structure-based models. $\fol{[NBF]}$, $\fol{[Sim]}$, $\fol{[RotE]}$ and $\fol{[Comp]}$ represents NBFNet, SimKGC, RotatE and CompleX models respectively. $\fol{[NBF]}$ does not scale to YAGO3-10. Best individual model results are underlined.}
\vspace{-0.05in}
\begin{tabular}{|p{2.7cm}|ccccc|ccccc|} 
\Xhline{3\arrayrulewidth} 
\multirow{2}{4em}{\textbf{Model}} & \multicolumn{5}{c|}{\textbf{WN18RR}} &
\multicolumn{5}{c|}{\textbf{FB15k-237}} \\
& MR & MRR & H@1 & H@3 & H@10 & MR & MRR & H@1 & H@3 & H@10 \\
\Xhline{3\arrayrulewidth}
 $\fol{[Sim]}$ &  \underline{174.0} & \underline{66.4} & \underline{58.5} & \underline{71.3} & \underline{80.3} & 131.9 & 32.1 & 23.2 & 34.6 & 50.5 \\
$\fol{[NBF]}$ & 699.3 & 54.2 & 48.6 & 56.9 & 65.7 & \underline{111.4} & \underline{40.5} & \underline{31.0} & \underline{44.3} & \underline{59.4}  \\
$\fol{[RotE]}$ & 4730.7 & 47.7 & 43.9 & 49.1& 55.2& 176.6 & 33.7 & 24.0 &37.4 & 53.2 \\
$\fol{[Comp]}$ & 5102.6 & 47.2 & 42.8 & 49.2 & 56.0& 180.7 & 35.7 & 26.3 & 39.4 & 54.7  \\
\cline{1-11}
$\fol{[Sim]}$+$\fol{[NBF]}$ & \textbf{56.6} & \textbf{73.2} & \textbf{66.9} & \textbf{76.5} & \textbf{85.7} & 92.2 & 42.7 & 33.2 & 46.7 & 61.5  \\
$\fol{[Sim]}$+$\fol{[RotE]}$ & 162.7 & 68.0 & 60.7 & 72.2 & 80.7 & 116.0 & 36.6 & 26.9 & 40.2 & 56.3 \\
$\fol{[Sim]}$+$\fol{[Comp]}$ & 172.9 & 68.0 & 60.8 & 72.3 & 80.7  & 116.3 & 37.8 & 28.3 & 41.2 & 57.1 \\
\cline{1-11}
$\fol{[Sim]}$+$\fol{[NBF]}$+$\fol{[RotE]}$  & \textbf{56.6} & \textbf{73.2} & \textbf{66.9} & \textbf{76.5} & \textbf{85.7} & \textbf{91.5} & \textbf{43.0} & \textbf{33.4} & \textbf{47.0} & \textbf{62.0} \\
$\fol{[Sim]}$+$\fol{[NBF]}$+$\fol{[Comp]}$ & \textbf{56.6} & \textbf{73.2} & \textbf{66.9} & \textbf{76.5} & \textbf{85.7} & 92.0 & 42.8 & 33.3 & 46.8 & 61.5  \\
\Xhline{3\arrayrulewidth}
\multirow{2}{4em}{\textbf{Model}}  & \multicolumn{5}{c|}{\textbf{CoDex-M}} & \multicolumn{5}{c|}{\textbf{Yago3-10}} \\
& MR & MRR & H@1 & H@3 & H@10 & MR & MRR & H@1 & H@3 & H@10 \\
\Xhline{3\arrayrulewidth}
 $\fol{[Sim]}$  & \underline{284.2} & 29.1 & 21.0  & 31.5 & 45.2 & \textbf{\underline{497.4}} & 15.8 & 10.0 & 16.2 & 27.3 \\
$\fol{[NBF]}$  & 337.5 & \underline{35.3} & 27.0 & \underline{39.0} & \underline{51.4} & -  & - & - & - & - \\
$\fol{[RotE]}$ & 502.6 & 33.5 & 26.3 & 36.8 & 46.9 & 1866.8 & \underline{49.3} & 39.9 & \underline{55.0} & \underline{67.1}  \\
$\fol{[Comp]}$ & 391.0 & \underline{35.3} & \underline{27.7} & 38.8 & 49.5 & 1578.1 & 49.2 & \underline{40.1} & 53.8 & 66.7 \\
\cline{1-11}
$\fol{[Sim]}$+$\fol{[NBF]}$ & 252.1 & 38.9 & 30.5 & 42.7 & \textbf{54.8} &  - & - & - & - & - \\
$\fol{[Sim]}$+$\fol{[RotE]}$ & 293.4 & 36.3 & 28.1 & 40.0 & 51.7 & 610.6 & \textbf{50.6} & \textbf{41.3} & \textbf{56.0}  & \textbf{67.9}   \\
$\fol{[Sim]}$+$\fol{[Comp]}$ & 296.3 & 37.5 & 29.6 & 41.0 & 52.4 & 515.9 & 49.5 & 40.5 & 54.2 & 66.6\\
\cline{1-11}
$\fol{[Sim]}$+$\fol{[NBF]}$+$\fol{[RotE]}$ & \textbf{216.5} & \textbf{40.0} & \textbf{31.2} & \textbf{43.3} & \textbf{54.8} & - & - & - & - & - \\
$\fol{[Sim]}$+$\fol{[NBF]}$+$\fol{[Comp]}$   & 293.3 & 37.6 & 29.8 & 41.1 & 52.5 & - & - & - & - & - \\
\Xhline{3\arrayrulewidth}
\end{tabular}
\label{tab:appendixMainDetailedresults}
\vspace{0.20in}
\centering
\small
\caption{Results of $\fol{[Sim]}$, $\fol{[NBF]}$, $\fol{[RotE]}$, $\fol{[Comp]}$ and $\fol{[Sim]}$ + $\fol{[NBF]}$ on the Reachable and Unreachable splits of WN18RR, FB15k-237, and CoDex-M datasets. Best individual model results are underlined.}

\vspace{-0.05in}
\begin{tabular}{|p{1.5cm}|p{1.5cm}|cccc|cccc|} 
\Xhline{3\arrayrulewidth} 
\multirow{2}{4em}{\textbf{Dataset}} & \multirow{2}{4em}{\textbf{Model}} & \multicolumn{4}{c|}{\textbf{Reachable Split}} &
\multicolumn{4}{c|}{\textbf{Unreachable Split}} \\
& & MR & MRR & H@1 &  H@10 & MR & MRR & H@1 &  H@10 \\
\Xhline{3\arrayrulewidth}
 \multirow{5}{4em}{\textbf{WN18RR}} & $\fol{[Sim]}$  & 29.7 & 85.3 & 79.4 & 94.5 & \underline{288.5} & \underline{51.8} & \underline{42.3} & \underline{69.0}  \\
 & $\fol{[NBF]}$ & \underline{4.7} & \underline{89.7} & \underline{86.8} & \underline{95.7} & 1250.7 & 26.0 & 18.3 & 41.8   \\
& $\fol{[RotE]}$ & 102.9 & 85.6 & 83.3 & 90.0 & 8404.8 & 17.5 & 12.6 & 1.1  \\
& $\fol{[Comp]}$ & 285.9 & 85.6 & 83.8 & 88.5 & 10526.6 & 16.2 & 12.1  & 23.7  \\
&  $\fol{[Sim]}$+$\fol{[NBF]}$ & \textbf{2.7} & \textbf{93.9} & \textbf{91.7} & \textbf{97.4} & \textbf{99.4} & \textbf{56.8} & \textbf{47.0} & \textbf{76.4} \\
\Xhline{3\arrayrulewidth}
 \multirow{5}{4em}{\textbf{FB15k}$-$\textbf{237}}  & $\fol{[Sim]}$ & 131.8 & 31.5 & 22.6 & 49.6 & \underline{153.8} & \underline{30.0} & 21.2 & \underline{48.2} \\
 & $\fol{[NBF]}$ & \underline{86.9} & \underline{44.8} & \underline{35.2} & \underline{64.0} & 180.4 & 28.2 & 19.3 & 46.2  \\
& $\fol{[RotE]}$ & 131.8 & 35.6 &  25.5 & 56.3 & 303.2 & 28.1 & 19.8  & 44.5 \\
& $\fol{[Comp]}$ & 129.9 & 37.9 & 28.0 & 57.8 & 323.9 & 29.7 & \underline{21.5} & 46.0 \\
& $\fol{[Sim]}$+$\fol{[NBF]}$ & \textbf{76.0} & \textbf{46.5} & \textbf{36.8} & \textbf{65.3} & \textbf{137.0} & \textbf{32.3} & \textbf{23.1} & \textbf{50.6} \\
\Xhline{3\arrayrulewidth}
 \multirow{5}{4em}{\textbf{CoDex}$-$\textbf{M}}  & $\fol{[Sim]}$ & 166.5 & 35.5 & 26.8 & 52.4 & \underline{363.6} & 23.7 & 15.8 & 39.6 \\
 & $\fol{[NBF]}$ & \underline{150.1} &\underline{47.8} & \underline{39.5} & \underline{63.2} & 458.1 & 27.2 & 18.9 & \underline{43.5} \\
& $\fol{[RotE]}$ & 290.5 & 44.2 & 37.2 & 56.8 & 639.1 & 26.7 & 19.5 & 40.5  \\
& $\fol{[Comp]}$ & 187.1 & 46.5 & 38.8 & 60.4 & 519.0 & \underline{28.2} & \underline{20.8} & 42.6  \\
& $\fol{[Sim]}$+$\fol{[NBF]}$ & \textbf{112.8} & \textbf{51.2} & \textbf{40.5} & \textbf{66.1} & \textbf{339.6} & \textbf{31.4} & \textbf{23.0}  & \textbf{47.6} \\
\Xhline{3\arrayrulewidth}
\end{tabular}
\label{tab:appendixReachabilityAblation}
\vspace{0.20in}
\end{table*} 

\section{Data Statistics and Evaluation Metrics}
\label{appendix: datastatistics}

Table \ref{tab:KG statistics} outlines the statistics of the datasets utilized in our experimental section. We utilize the standard train, validation and test splits for all datasets.
\paragraph{Metrics:} For each triplet $\fol{(h, r, t)}$ in the KG, typically queries of the form $\fol{(h, r, ?)}$ and $\fol{(?, r, t)}$ are created for evaluation, with corresponding answers $\fol{t}$ and $\fol{h}$. We represent the $\fol{(?, r, t)}$ query as $\fol{(t, r^{-1}, ?)}$ with the same answer $\fol{h}$, where $\fol{r^{-1}}$ is the inverse relation for $\fol{r}$, for both training and testing. Given ranks for all queries, we report the Mean Reciprocal Rank (MRR) and Hit@k (H@k, k = 1, 10) under the filtered setting in the main paper and two additional metrics: Mean Rank (MR) and Hits@3 in the appendices.

\section{Detailed Results on Proposed Ensemble}
\label{appendix:mainresultsindetail}
Here we present our experimental setup for the main results presented in Table \ref{tab:tablemainMRRAndHitsAt10}. Since loading both $\fol{NBFNet}$ and the two BERT encoders from $\fol{SimKGC}$ into GPU at the same time is too taxing for our hardware, we dump the embeddings of all possible $\fol{(h,r)}$ and $\fol{t}$ from $\fol{SimKGC}$ to disk, and use them for training our ensemble. $\fol{SimKGC}$ is reliant on textual descriptions for performance. The original authors provide descriptions for WN18RR and FB15k-237, while descriptions for CoDex-M are available as part of the dataset. Since YAGO3-10 does not contain any descriptions, we treat the entity names as their descriptions. $\fol{SimKGC}$ also has a structural re-ranking step independent of its biencoder architecture, which we do not utilize as we expect our ensembling method to subsume it.

Next, we present results in Table \ref{tab:appendixMainDetailedresults}  that are supplementary to results already presented in Table \ref{tab:tablemainMRRAndHitsAt10}. In addition to MRR, Hits@1 and Hits@10 considered in Table \ref{tab:tablemainMRRAndHitsAt10}, we also present numbers for Mean Rank (MR) and Hits@3 in Table \ref{tab:appendixMainDetailedresults}. As before, the '$\fol{+}$' sign represents our ensemble approach. We also consider an additional KG embedding model ComplEx~\cite{complex2016} ($\fol{[Comp]}$ in tables) in this section and present complete results for it.

We observe that for the two new metrics considered in Table \ref{tab:appendixMainDetailedresults}, we also obtain substantial performance gains on ensembling, notably a gain of 5.2 pt Hits@3 and 67.4\% MR on the WN18RR dataset with $\fol{[Sim]}$ + $\fol{[NBF]}$ over $\fol{[Sim]}$. Further, we observe that $\fol{[Sim]}$+$\fol{[Comp]}$ consistently outperforms both $\fol{[Sim]}$ and $\fol{[Comp]}$, (by up to 2.2 pt MRR for CoDex-M). We also present complete numbers for $\fol{[Sim]}$+$\fol{[NBF]}$+$\fol{[Comp]}$ and $\fol{[Sim]}$+$\fol{[NBF]}$+$\fol{[RotE]}$ here. 

\section{Feature Selection for Ensemble Weight Learning}
\label{appendix:featureselection}
\begin{figure}[h]
\centering
\begin{center}
\begin{subfigure}{.5\textwidth}
  \centering
  \includegraphics[width=.8\linewidth]{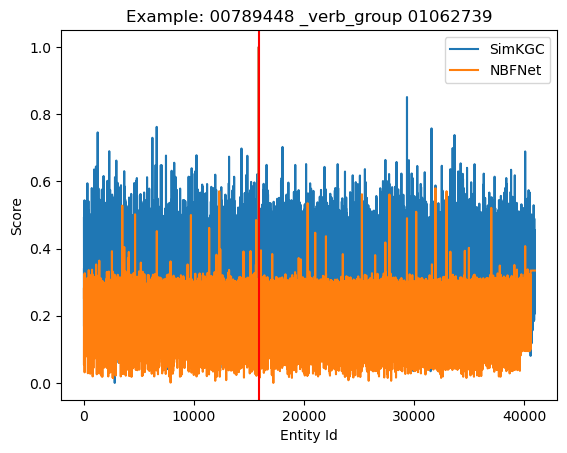}
  \caption{Query in reachable split}
  \label{fig:sub1}
\end{subfigure}
\begin{subfigure}{.5\textwidth}
  \centering
  \includegraphics[width=.8\linewidth]{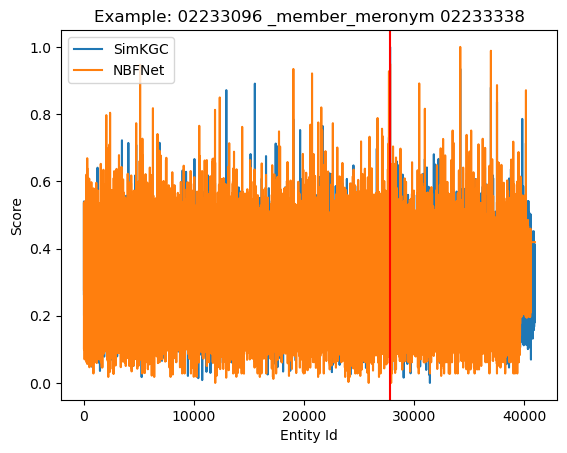}
  \caption{Query in unreachable split}
  \label{fig:sub2}
\end{subfigure}
\caption{Score distributions of $\fol{[NBF]}$ and $\fol{[Sim]}$ for two queries in WN18RR} 
\label{fig:test}
\end{center}
\end{figure}

In this section, we justify our choice of features for learning ensemble weights. We focus on $\fol{NBF}$ and $\fol{Sim}$ for this purpose. We claim that after our normalization procedure, a model has lower mean and variance when it is confident about the validity of its top predictions. To highlight this, we present distribution of the normalized scores over all candidate entities for $\fol{NBF}$ and $\fol{Sim}$ for two queries in the WN18RR dataset: one from the reachable split and the other from the unreachable split. The query for Figure \ref{fig:sub1} lies in the reachable split while the query for Figure \ref{fig:sub2} lies in the unreachable split. The entity id of the gold answer is marked with a red vertical line in both cases.

We notice that for the query in the reachable split, $\fol{[NBF]}$ is very confident about its top prediction. Therefore, it scores the gold answer significantly higher than the other candidates. Upon normalization, this causes the other entities to have comparatively smaller values (mostly in the range [0-0.4]), with a tighter spread. In comparison, for the query in the unreachable split, $\fol{[NBF]}$ cannot predict the gold answer confidently. This results in a much larger spread of scores across entities, with a lot of extreme values close to 1 indicating that the model is unable to conclusively determine which entity is the correct one. We choose the mean and variance as features because they will be able to distinguish between these two distributions, with their values being substantially smaller in the first case where $\fol{[NBF]}$ is confident about the predictions. 

$\fol{[Sim]}$ also has these properties, albeit to a lesser degree. This is because SimKGC cannot exploit the KG structure, and therefore has to draw conclusions based on textual descriptions, which can point to several candidate answers of seemingly comparable validity. This results in the score distributions having a higher spread and a lower margin between the score of the top prediction and the other candidates. Therefore, the relative values of these mean and variance features can also inform the MLPs about the relative confidence of the models about their output, allowing them to compute ensemble weights for corresponding models as necessary. 

As validation, we present the average of the mean and variance features from $\fol{[NBF]}$ over all test queries in the reachable and unreachable split for the WN18RR, FB15k-237 and CoDex-M datasets in Table \ref{tab:analysisfeat}. 

\begin{table}[h]
\caption{Average of$\fol{[NBF]}$ Features across Splits }
\centering
\begin{center}
\resizebox{\columnwidth}{!}{
\begin{tabular}{|p{2.5cm}| p{1.5cm}p{2cm}p{1.5cm}p{2cm}|} 
 \cline{1-5}
\multirow{2}{4em}{\textbf{Dataset}} & \multicolumn{2}{c}{\textbf{Reachable Split}}  & \multicolumn{2}{c|}{\textbf{Unreachable Split}} \\
 & \textbf{Mean} & \textbf{Var} & \textbf{Mean} & \textbf{Var} \\
 \cline{1-5}
\textbf{WN18RR} & 0.277 & 0.008 & 0.353 & 0.127 \\
 \textbf{FB15k-237} & 0.244 & 0.017 & 0.284 & 0.019 \\
 \textbf{CoDex-M} & 0.492 & 0.015 & 0.566 & 0.017 \\
 \cline{1-5}
\end{tabular}
}
\end{center}
\vspace{-0.20in}
\label{tab:analysisfeat}
\end{table}

We find that the average of the mean and variance features is up to 21\% lower (for WN18RR) on the reachable split than the unreachable split, allowing the MLPs to distinguish between the splits based on score distribution statistics alone. We finally present results of experiments with other similar sets of features as input to the MLP in Table \ref{tab:feat}.

\begin{table}[H]
 \caption{Performance of $\fol{[Sim]}$ + $\fol{[NBF]}$ with different sets of input features to the MLP. + indicates concatenation here. Var and Std stand for variance and standard deviation. Zip stands for passing the entire output distribution from the base models as input to the MLP. Top 10 indicates using the top 10 scores from the output distribution as input features.}
  \vspace{-0.07in}
 \label{tab:feat}
 \centering
\small
\begin{center} 
\resizebox{\columnwidth}{!}{
\begin{tabular}{|p{1.8cm}|c|ccc|} 
\Xhline{3\arrayrulewidth}
\textbf{Dataset} & \textbf{Input Features} & MRR & H@1 & H@10 \\
\cline{1-5}
 \multirow{6}{4em}{\textbf{WN18RR}}& Mean + Var & \textbf{73.2} & \textbf{66.9} & \textbf{85.7} \\
 & Mean + Std & 73.1 & 66.8 & 85.1 \\
 & Std & 67.1 & 59.2 & 80.5 \\
 & Mean & 67.2 & 59.4 & 80.5 \\
 & Zip & 66.6 & 58.7 & 80.3 \\
 & Top 10 & 72.1 & 65.5 & 84.8 \\
\cline{1-5}
 \multirow{6}{5em}{\textbf{CoDex-M}}& Mean + Var & \textbf{38.9} & \textbf{30.5} & \textbf{54.8} \\
 & Mean + Std & 38.1 & 29.9 & 54.1 \\
 & Std & 32.5 & 23.4 & 48.5 \\
 & Mean & 32.1 & 23.5 & 48.6 \\
 & Zip & 31.6 & 23.3 & 48.1 \\
 & Top 10 & 31.7 & 23.3 & 48.4 \\
\cline{1-5}
\end{tabular}
}
\end{center}
\vspace{-0.15in}
\end{table}

We find that features that are created according to the reasoning above (Mean + Var and Mean + Std) perform better as compared to other features (Zip, Top 10) across datasets and metrics.


\section{Choice of Training Data for Dynamic Ensemble}
\label{appendix:choice}
In this section, we expand upon the choice of using the validation split to train the dynamic ensemble weights, which is usually used for manually tuning the constant ensemble weights in static ensembling. We present results for dynamic ensembles trained on three splits of data: i) the full training split ($\fol{Full Train}$) ii) the validation split ($\fol{Validation}$, this corresponds to the dynamic ensemble results in the paper) iii) a randomly-chosen 1\% split of the training data, which is held-out while training the base models before ensembling ($\fol{Held-Out Train}$). We present results for $\fol{[Sim]} + \fol{[NBF]})$ trained on these three splits of the WN18RR and FB15k-237 datasets in Table \ref{}. 

\begin{table}[th]
 \caption{Results for $\fol{[Sim]} + \fol{[NBF]}$ trained under the $\fol{Full Train}$, $\fol{Validation}$ and $\fol{Held-Out Train}$ conditions on the WN18RR and FB15k-237 datasets. $\fol{BestIndv}$ represents the performance of the best individual model in each case, which is $\fol{[Sim]}$ for the WN18RR dataset and $\fol{[NBF]}$ for the FB15k-237 dataset.}
  \vspace{-0.07in}
 \label{tab:ModelComparison}
 \centering
\small
\begin{center} 
\resizebox{\columnwidth}{!}{
\begin{tabular}{|c|ccc|ccc|} 
\Xhline{3\arrayrulewidth}
\multirow{2}{4em}{\textbf{Method}} & 
\multicolumn{3}{c|}{\textbf{WN18RR}} &
\multicolumn{3}{c|}{\textbf{FB15k-237}} \\
 & MRR & H@1 & H@10 & MRR & H@1 & H@10 \\
\Xhline{3\arrayrulewidth}
  $\fol{BestIndv}$  & 66.4 & 58.5 & 80.3 & 40.5 & 31.0 & 59.4\\
  $\fol{Full Train}$  & 66.4 & 58.6 & 80.3 & 40.6 & 31.2 & 59.4\\
  $\fol{Validation}$  & \textbf{73.2} & \textbf{66.9} & \textbf{85.7} & \textbf{42.7} & \textbf{33.2} & \textbf{61.5}\\
  $\fol{Held-Out Train}$  & 73.0 & 66.5 & 85.4 & 42.4 & 32.9 & 61.4\\
  \Xhline{3\arrayrulewidth}
\end{tabular}
}
\end{center}
\vspace{-0.15in}
\end{table}

We find that $\fol{Full Train}$ results in less than 0.1 pt MRR improvement over best individual models for both datasets. This is because both base models are capable of fitting the training data with near-perfect performance. As a result, both models showcase high confidence about their outputs and the dynamic ensemble is unable to learn any correlations between model confidence and corresponding ensemble weight for the test split. Therefore, the ensemble weights for each model converge rapidly to 0 or 1 during training.

 We additionally find that $\fol{Held-Out Train}$ results in performance within 0.3 pt MRR of $\fol{Validation}$ in both datasets. This small drop in performance might be caused by the slightly smaller amount of data being used to train both the base models and the dynamic ensemble, as compared to the original setting. This shows that holding out part of the training data is an effective strategy to train the dynamic ensemble on datasets that do not have a validation split, as the small drop in performance of the base model is amply compensated by the gains from ensembling.

\section{Detailed Reachability Ablation}
\label{appendix:detailedablation}
In this section we discuss further results of the experiment done to answer \textbf{Q1} in Section \ref{sec:analysis}. The results presented in Table \ref{tab:appendixReachabilityAblation} are supplementary to the results presented in Table \ref{tab:ReachabilityAblation} where in addition to the MRR, Hits@1, Hits@10 metrics already presented in Table \ref{tab:ReachabilityAblation}, we present results over one additional metric, MR. Additionally, we present the results on the `reachable' and `unreachable' split of CoDex-M, and for $\fol{[RotE]}$ and $\fol{[Comp]}$ on all datasets. We observe that $\fol{[Sim]}$ has up to 76\% lower MR than $\fol{[NBF]}$ on the unreachable split while $\fol{[NBF]}$ has up to 83.3\% lower MR than $\fol{[Sim]}$ on the reachable split over all the three datasets (both statistics mentioned are for WN18RR). 
The ensemble of $\fol{[Sim]}$+$\fol{[NBF]}$ brings the MR down further, notably obtaining a gain of 42.5\% on reachable split and 66\% on unreachable split over best individual models for the WN18RR dataset. We also observe that $\fol{[RotE]}$ and $\fol{[Comp]}$ show similar variation of performance across splits when compared to $\fol{[NBF]}$, performing notably better on the reachable split as compared to the unreachable split across datasets. This indicates that these KG embedding models are also dependent on KG structure and paths between the head and gold tail to some extent for performance. We investigate this in more detail in Appendix \ref{appendix:rotatestructure}. 
 
\section{RotatE as a Structure-Based Model}
\label{appendix:rotatestructure}
We claim that despite structure not being explicitly involved in the training of $\fol{[RotE]}$, it is still capable of capturing the structure of the KG to some extent in its relation embeddings by exploiting the compositionality inherent in its scoring function. Consider an example in which $\fol{(h_1, r_1, h_2)}$, $\fol{(h_2, r_2, h_3)}$ and $\fol{(h_1, r_3, h_3)}$ are all present in the KG. Let $\fol{T_{r}(h)}$ be the vector obtained after rotating the embedding of $\fol{h}$ by the complex rotation defined by $\fol{r}$. During training, $\fol{T_{r_1}(h_1)}$ will be brought close to the embedding of $\fol{h_2}$ and $\fol{T_{r_2}(h_2)}$ will be brought close to the embedding of $\fol{h_3}$. As a result, $\fol{T_{r_2}(T_{r_1}(h_1))}$ will be brought close to $\fol{h_3}$. Upon training on $\fol{(h_1, r_3, h_3)}$, $\fol{T_{r_3}(h_1))}$ will also be brought close to $\fol{h_3}$. However, the relative positions of $\fol{h_1}$ and $\fol{h_3}$ on the complex plane already contain information about $\fol{T_{r_2} \circ T_{r_1}}$, which is used while training $\fol{T_{r_3}}$. As more such examples are seen over multiple epochs, $\fol{T_{r_3}}$ will eventually be brought closer to the composed rotation $\fol{T_{r_2} \circ T_{r_1}}$. Therefore, when query $\fol{(h, r_3, ?)}$ is seen at test time, the model will be more likely to return candidates $\fol{t}$ which are connected to $\fol{h}$ through a path in the KG involving relations $\fol{r_1}$ and $\fol{r_2}$, making it structure dependent.

Of course, this phenomenon is not limited to paths of length 2, but can encode paths of longer length as well. We also expect only the most common paths to be captured through this mechanism, since multiple such paths have to be encoded by the same relation embedding. To validate these claims, we perform an experiment where we exhaustively mine the dataset for cases where $\fol{(h_1, r_1, h_2)}$ is present in the KG, alongside an entity $\fol{h_3}$ such that $\fol{(h_1, r_2, h_3)}$ and $\fol{(h_3, r_3, h_2)}$ are also present in the KG. This essentially considers all the cases where there is a path involving $\fol{r_2}$ and $\fol{r_3}$ that is closed by $\fol{r_1}$. We enumerate all such cases for each triple $\fol{(r_1, r_2, r_3)}$ and filter out those triples that have less than 20 occurrences in the KG. For each of the remaining triples, we take a random vector and transform it according to $\fol{T_{r_2} \circ T_{r_3}}$. We then report the $\fol{r}$ such that transforming the same vector according to $\fol{T_{r}}$ moves it closest to the result obtained on transforming it according to $\fol{T_{r_2} \circ T_{r_3}}$. We expect $\fol{r}$ to be $\fol{r_1}$ for a majority of the triples based on our claims. We report accuracies obtained through this experiment for $\fol{[RotE]}$ on the WN18RR, FB15k-237 and CoDex-M datasets in Table \ref{tab:appendixpathdependencyinRotatE}. 

\begin{table}[H]
 \caption{Structure Dependency of $\fol{[RotE]}$ and $\fol{[Comp]}$ }
  \vspace{-0.07in}
 \label{tab:appendixpathdependencyinRotatE}
 \centering
\small
\begin{center} 
\resizebox{\columnwidth}{!}{
\begin{tabular}{|c|c|} 
\Xhline{3\arrayrulewidth}
\multirow{2}{4em}{\textbf{Dataset}} &
\textbf{Accuracy of Closest Relation} \\
 & $\fol{[RotE]}$ \\
\Xhline{3\arrayrulewidth}
\textbf{WN18RR}&  38.1 \\
\textbf{FB15k-237} & 45.0  \\
\textbf{CoDex-M} & 68.2  \\
\Xhline{3\arrayrulewidth}
\end{tabular}
}
\end{center}
\vspace{-0.20in}
\end{table}
\begin{table*}[t]
\centering
\small
\caption{Results of paired student's t-test for dynamic ensemble and static ensemble on MRR with $\fol{[Sim]}$ + $\fol{[NBF]}$.}

\vspace{-0.05in}
\begin{tabular}{|p{1.5cm}|p{2.5cm}|ccccc|} 
\Xhline{3\arrayrulewidth} 
Dataset & Method & Split 1 & Split 2 & Split 3 & Split 4 & Split 5 \\
\Xhline{3\arrayrulewidth}
 \multirow{3}{5em}{\textbf{WN18RR}} & Dynamic Ensemble  & 73.5 & 73.2 & 73.2 & 73.7 & 73.2 \\
 & Static Ensemble  & 72.0 & 72.5 & 71.9 & 72.3 & 71.9   \\
& Difference & 1.5 & 0.7 & 1.3 & 1.4 & 1.3  \\ \hline
 \multirow{3}{5em}{\textbf{CoDex-M}} & Dynamic Ensemble  & 38.7 & 38.9 & 38.8 & 39.1 & 38.7 \\
 & Static Ensemble  & 37.1 & 37.8 & 38.0 & 37.8 & 38.0   \\
& Difference & 1.6 & 1.1 & 0.8 & 1.3 & 0.7  \\
\Xhline{3\arrayrulewidth}
\end{tabular}
\label{tab:paired}
\vspace{0.20in}
\end{table*}

We find that the accuracies are substantially better than the random baseline of $\frac{\fol{1}}{\fol{Number of Relations}}$ for all datasets (which is 9.1\% for WN18RR, 0.4\% for FB15k-237 and 1.4\% for CoDex-M). $\fol{[Sim]}$ is not capable of capturing this notion, since it encodes $\fol{(h, r)}$ together using BERT, not as a composition of $\fol{h}$ and $\fol{r}$ embeddings. Therefore, we find that its behavior is independent of the split in which the query under consideration is present. 


\section{Reachability Trends of Ensemble Weights}
\label{appendix:detailedweights}
The aim of this section is to further discuss the results of the experiment done to answer \textbf{Q2} in Section \ref{sec:analysis}. The results in Table \ref{tab:appendixanalysisofensembleweights} present the mean and standard deviation of ensemble weights $\fol{w_2}$ over the queries in the reachable and unreachable split for the WN18RR, CoDex-M and FB15k-237 datasets. The weight discussed in these tables is $\fol{w_2}$ in the ensemble defined as $\fol{w_1[Sim]}$ + $\fol{w_2[NBF]}$ (with $\fol{w_1} = 1$) according to Section \ref{sec:main}. We observe that across all datasets, the average weight for reachable split is higher than the weight of unreachable split (up to 17\% higher for WN18RR), thus reinforcing the fact that our approach gives more weightage to $\fol{[NBF]}$ on the reachable split across datasets. The standard deviation of $\fol{w_2}$ is also non-trivial on all splits of all datasets, showing that our approach is capable of adjusting it as required by individual queries.

\begin{table}[H]
\caption{Mean and Standard Deviation (Std Dev in Table) of Ensemble Weights for $\fol{[Sim]+[NBF]}$ }
\vspace{-0.1in}
\centering
\small
\begin{center}
\resizebox{\columnwidth}{!}{
\begin{tabular}{|p{1.5cm}| p{1.5cm}p{1.5cm}p{1.5cm}p{1.5cm}|} 
 \cline{1-5}
\multirow{2}{4em}{\textbf{Dataset}} & \multicolumn{2}{c}{\textbf{Reachable Split}}  & \multicolumn{2}{c|}{\textbf{Unreachable Split}} \\
 & \textbf{Mean} & \textbf{Std Dev} & \textbf{Mean} & \textbf{Std Dev} \\
 \cline{1-5}
\textbf{WN18RR} & 0.61 & 0.04 & 0.52 & 0.07 \\
 \textbf{CoDex-M} & 2.03 & 0.24 & 1.91 & 0.38 \\
 \textbf{FB15k-237} & 2.64 & 0.22 & 2.57 & 0.24 \\
 \cline{1-5}
\end{tabular}
}
\end{center}
\vspace{-0.20in}
\label{tab:appendixanalysisofensembleweights}
\end{table}

To investigate why our ensemble weights are not binary and are quite consistent with each other for each split, we contrast $\fol{[Sim]}$ + $\fol{[NBF]}$ with a model that selects NBFNet on the reachable split and SimKGC on the unreachable split: $\fol{Split-Select}$. We present results in Table \ref{tab:split}.

\begin{table}[H]
 \caption{Comparison of $\fol{[Sim]}$ + $\fol{[NBF]}$ and $\fol{Split-Select}$ on the WN18RR dataset.}
  \vspace{-0.07in}
 \label{tab:split}
 \centering
\small
\begin{center} 
\resizebox{\columnwidth}{!}{
\begin{tabular}{|p{1.5cm}|c|ccc|} 
\Xhline{3\arrayrulewidth}
\textbf{Dataset} & \textbf{Approach} & MRR & H@1 & H@10 \\
\cline{1-5}
 \multirow{2}{4em}{\textbf{WN18RR}}& $\fol{[Sim]}$ + $\fol{[NBF]}$ & \textbf{73.2} & \textbf{66.9} & \textbf{85.7} \\
 & $\fol{Split-Select}$ & 68.4 & 61.8 & 81.0 \\
\cline{1-5}
\end{tabular}
}
\end{center}
\vspace{-0.15in}
\end{table}

We find that dynamic ensembling performs better than the oracle by 4.8 pt MRR. This is because structure-based models tend to rank more connected tails higher, while text-based models rank tails based solely on their textual descriptions. Therefore, a soft ensemble can take advantage of both structural and textual information to perform better than a mixture-of-experts model that simply selects one of the base models based on expected performance trends.  

\section{Significance of Improvements with Dynamic Ensembling}
\label{appendix:sig}

We first perform a paired student’s t-test on the MRR over a 5-fold split for $\fol{[Sim]}$ + $\fol{[NBF]}$ to confirm that the gains obtained by our approach over static ensembling are statistically significant. We present the results in Table \ref{tab:paired}.

We obtain a t-value of 8.9 for WN18RR and 6.7 for CoDex-M. With a p-value of 0.05, the reference value is 2.78. Therefore, the gains obtained by our model over static ensembling are indeed statistically significant. 

The performance of an ensemble is ultimately dependent on the performance of the individual models. To obtain an estimate of the best possible performance that can be obtained from model fusion, we present results in Table \ref{tab:best} with $\fol{[Sim]}$ + $\fol{[NBF]}$ for a model that selects the most performant model for each query ($\fol{BEST}$).

\begin{table}[H]
 \caption{Comparison of $\fol{[Sim]}$ + $\fol{[NBF]}$ and $\fol{BEST}$ on the WN18RR and CoDex-M datasets.}
  \vspace{-0.07in}
 \label{tab:best}
 \centering
\small
\begin{center} 
\resizebox{\columnwidth}{!}{
\begin{tabular}{|p{1.5cm}|c|ccc|} 
\Xhline{3\arrayrulewidth}
\textbf{Dataset} & \textbf{Approach} & MRR & H@1 & H@10 \\
\cline{1-5}
 \multirow{2}{4em}{\textbf{WN18RR}}& $\fol{[Sim]}$ + $\fol{[NBF]}$ & 73.2 & 66.9 & 85.7 \\
 & $\fol{BEST}$ & \textbf{74.1} & \textbf{67.6} & \textbf{86.1} \\
\cline{1-5}
 \multirow{2}{4em}{\textbf{CoDex-M}}& $\fol{[Sim]}$ + $\fol{[NBF]}$ & 38.9 & 30.5 & 54.8 \\
 & $\fol{BEST}$ & \textbf{41.2} & \textbf{32.6} & \textbf{57.9} \\
\cline{1-5}
\end{tabular}
}
\end{center}
\vspace{-0.15in}
\end{table}

We find that the results for our dynamic ensemble are only up to 2.3 MRR pts behind a theoretical oracle that always knows the best model for each query, indicating that most of the potential for improvement through late fusion techniques has been obtained through dynamic ensembling.



\end{document}